\documentclass[lettersize,journal]{IEEEtran}
\usepackage{amsmath,amsfonts}
\usepackage{algorithmic}
\usepackage{algorithm}
\usepackage{array}
\usepackage[caption=false,font=normalsize,labelfont=sf,textfont=sf]{subfig}
\usepackage{textcomp}
\usepackage{stfloats}
\usepackage{url}
\usepackage{verbatim}
\usepackage{graphicx}
\usepackage{cite}
\usepackage{cite}
\usepackage{amsmath,amssymb,amsfonts}
\usepackage{algorithmic}
\usepackage{textcomp}

\usepackage{graphicx}
\graphicspath{{.}{./Image/}}

\usepackage{pifont}
\newcommand{\cmark}{\text{\ding{51}}}
\newcommand{\xmark}{\text{\ding{55}}}

\usepackage{booktabs}

\usepackage{bbm}
\usepackage{url}

\usepackage{listings}
\usepackage{xcolor}
\usepackage{xspace}
\newcommand{\revise}[1]{{\color{black} #1}\xspace}

\begin{document}

\title{Client Selection in Federated Learning: Principles, Challenges, and Opportunities}

\author{
Lei Fu, Huanle Zhang, Ge Gao, Mi Zhang~\IEEEmembership{Senior Member, IEEE}, and Xin Liu~\IEEEmembership{Fellow, IEEE}
\IEEEcompsocitemizethanks{\IEEEcompsocthanksitem 
Lei Fu is with the Bank of Jiangsu and Fudan University, China. 
E-mail: leileifu@163.sufe.edu.cn;
Huanle Zhang and Ge Gao are with the School of Computer Science and Technology, Shandong University, China.
E-mail: \{dtczhang, vivi\_gaoge\}@sdu.edu.cn;
Mi Zhang is with the Department of Computer Science and Engineering, Ohio State University, USA. 
E-mail: mizhang.1@osu.edu;
Xin Liu is with the Department of Computer Science, University of California, Davis, USA. 
E-mail: xinliu@ucdavis.edu.
}
\thanks{Huanle Zhang is the corresponding author. }
}

\maketitle

\begin{abstract}
As a  privacy-preserving paradigm for training  Machine Learning (ML) models, Federated Learning (FL) has received tremendous attention from both industry and academia. In a typical FL scenario,  clients exhibit significant heterogeneity in terms of data distribution and hardware configurations. Thus, randomly sampling clients in each training round may not fully exploit the local updates from heterogeneous clients, resulting in lower model accuracy, slower convergence rate, degraded fairness, etc. To tackle the FL client heterogeneity problem, various client selection algorithms have been developed, showing promising performance improvement. In this paper, we systematically present recent advances in the emerging field of FL client selection and its challenges and research opportunities.  
We hope to facilitate practitioners in choosing the most suitable client selection mechanisms for their applications, as well as inspire researchers and newcomers to better understand this exciting research topic.

\end{abstract}



\section{Introduction}
\label{sec:introduction}

\IEEEPARstart{F}{ederated} Learning (FL) has gained great momentum in both academia and industry~\cite{iotfl, fl20dsml}. As a decentralized paradigm that preserves data privacy while enabling Machine Learning (ML) model training, FL has been applied to various fields, 
such as next-word prediction~\cite{gboard}, financial fraud detection~\cite{financialfraud}, and healthcare data analysis~\cite{healthcareanalysis}. According to the POLARIS report, the global FL market was valued at USD 110.8 million in 2021 and is expected to grow at a compound annual growth rate of 10.7\%, reaching USD 266.8 million by 2030~\cite{flmarket}.

FL models generally require a preliminary training process which, typically, includes four main steps, as illustrated in Figure 1.
The training steps repeat until the global model on the server converges or a predefined number of epochs is reached. 
A plethora of work aims to solve different aspects of FL training, such as optimized aggregation methods~\cite{fedNova20neurips, fedyogi}, enhanced privacy protection~\cite{privacy22access, privacyblockchain22tii}, and improved robustness~\cite{fl-wbc21neurips, ditto21icml}. 

\begin{figure}[!t]
    \centering
    \includegraphics[width=\columnwidth]{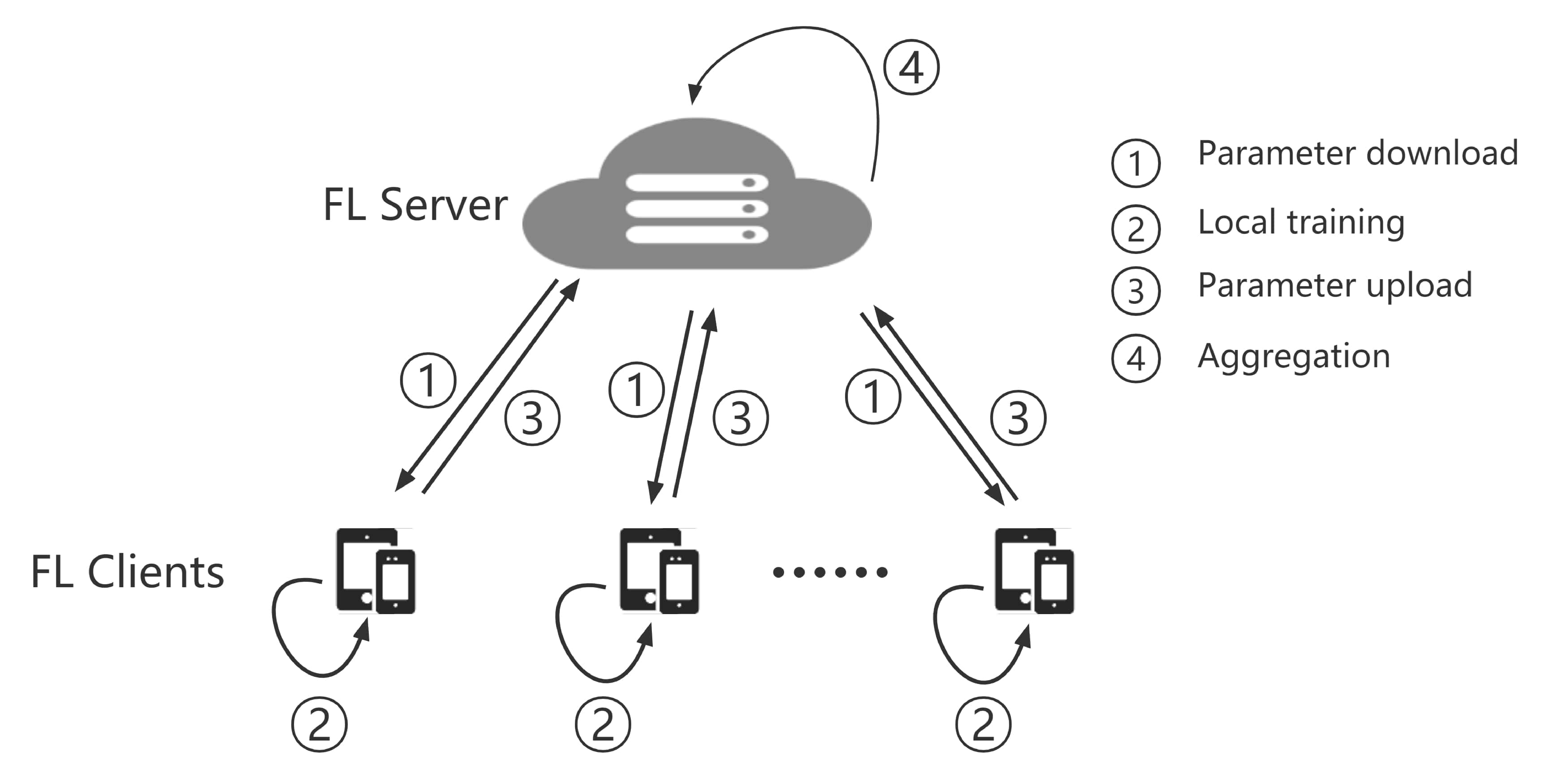}
    \caption{An illustration of a  standard FL training process that incorporates four steps: model parameter download from the server to the clients, local training on the clients, model parameter upload from the clients to the server, and model aggregation on the server.}
    \label{fig:fl_overview}
\end{figure}

\begin{table*}[!t]
    \centering
    \begin{tabular}{c c | c c | p{3.5in}}
        Work & Publication & System & Data & Description \\
          \toprule
FedCS~\cite{fedcs19icc} & ICC'19 & \cmark & \xmark & Select as many clients as possible within a specified deadline. \revise{It is based on a greedy algorithm with a knapsack constraint.} \\
Elsa et al.~\cite{norm21icassp} & ICASSP'21 & \xmark & \cmark & Two-level importance sampling for clients and data. \revise{It first selects clients and then samples data of the selected clients.} \\
OCEAN~\cite{ocean21twc} & TWC'21 & \cmark & \xmark & Bandwidth allocation under client energy constraints. \revise{It utilizes wireless channel information to achieve a better client selection pattern.}\\
Oort~\cite{oort21osdi} & OSDI'21 & \cmark & \cmark & Exploit data and system heterogeneity in clients. \revise{It employs an exploration-exploitation strategy to select participants for robustness to outliers.} \\
Mohammad et al. \cite{wireless21twc} & TWC'21 & \cmark & \cmark & Consider shared block fading wireless channels and local gradient norm. \revise{It designs a resource allocation policy to schedule low-profile client devices.} \\
FOLB~\cite{folb21jsac} & JSAC'21 & \cmark & \cmark & Based on the correlation between local update and global update. \revise{It estimates clients' capabilities of contribution to the updates.}  \\
Wenlin et al.~\cite{norm20arxiv} & \revise{TMLR'22} & \xmark & \cmark & Norm-based client selection to tackle the communication bottleneck. \revise{It approximates the optimal formula for client selection with a simple algorithm.} \\  
FedPNS~\cite{fedpns22tnse} & TNSE'22 & \xmark & \cmark & Removing adverse local updates by comparing the gradients of the local and the global. \revise{It preferentially selects clients that propel faster model convergency.}  \\
POWER-OF-CHOICE~\cite{powerofchoice22aistats} & AISTATS'22 & \cmark & \cmark & Local loss based client selection tradeoff between convergence speed and solution bias. \revise{It shows that biasing client selection can speed up the convergence.} \\ 
PyramidFL~\cite{pyramidfl22mobicom} & MobiCom'22 & \cmark & \cmark & Exploit data and system heterogeneity within selected clients. \revise{It determines the utility-based client selection and then optimizes utility profiling locally.}\\
FCFL~\cite{fcfl22imwut} & IMWUT'22 & \cmark & \cmark & Wearable devices in inferior networking conditions. \revise{It proposes movement aware FL to aggregate only the model udpates with top contributions.} \\
Eiffel~\cite{eiffel22tpds} & TPDS'22 & \cmark & \cmark & Jointly consider factors such as resource demand and the age of update. \revise{It adaptively adjust the frequency of local and global model updates.} \\
Bing et al.~\cite{fl22infocom} & INFOCOM'22 & \cmark & \cmark & Optimization of client sampling probabilities to address system and statistical heterogeneity. \revise{It minimizes the wall-clock convergence time.} \\
\revise{F3AST~\cite{f3ast23jstsp}} & \revise{JSTSP'23} & \revise{\cmark} & \revise{\xmark} & \hspace{0.0001in}\revise{Learns an availability-dependent client selection strategy to minimize the impact of client-selection variance on the global model's convergence.} \\
\revise{Haoyu et al.~\cite{fl23iotj}} & \revise{IoT-J'23} & \revise{\cmark} & \revise{\xmark} & \hspace{0.0001in}\revise{Propose a dynamic user and task scheduling scheme with a block-wise incremental gradient aggregation algorithm.} \\
        \bottomrule
    \end{tabular}
   \vspace{0.05in} \caption{Representative work of FL client selection algorithms, ordered by the publication year. We only list work with explicit utility measurement and scheduling decisions for clients. We tag whether the work designed for system heterogeneity and data heterogeneity.}
    \label{tab:papers}
\end{table*}

Based on the scale and approach of training, FL can be roughly categorized into cross-silo FL and cross-device FL~\cite{flField}. Cross-silo FL targets collaborative learning among several organizations, while cross-device FL targets machine learning across large populations, e.g., mobile devices~\cite{flField, fdatscale19sysml}. 
Currently, cross-device FL is more widely used in various application domains such as mobile phones, Internet-of-Things (IoT)~\cite{ioft}, and mobile edge computing~\cite{adaptiveMecFl19jsac}. In cross-device scenarios, FL clients (e.g., all sorts of mobile or IoT devices) exhibit significant heterogeneity in terms of data statistics and system configurations, which, if not handled appropriately, can degrade the FL performance~\cite{flash22tmc}.

Thus, to solve client heterogeneity problems, FL client selection (a.k.a. participant selection or device sampling) is an emerging topic. FL client selection decides which client devices are chosen in each training round.
An effective FL client selection scheme can significantly improve model accuracy~\cite{oort21osdi}, enhance fairness~\cite{eiffel22tpds}, strengthen robustness~\cite{folb21jsac}, and reduce training overheads~\cite{pyramidfl22mobicom}. Therefore, the research community is witnessing a rapid development of FL client selection research in recent years~\cite{flClientSelectionSurvey}.

However, there lacks a high-quality overview paper on FL client selection that can help newcomers quickly acquaint themselves with this research topic. To fill this gap, we provide an overview of FL client selection, covering the most representative work. Instead of simply listing existing work, we adopt a more systematic way: we organize existing papers based on their criteria for prioritizing FL clients. In addition,
we discuss challenges and research opportunities for FL client selection. 
To the best of our knowledge, this is the first overview paper that provides deep insights into FL client selection.

This paper is organized as follows. Section \ref{sec:slr} explains our literature review process. Section~\ref{sec:heterogeneity} clarifies the client heterogeneity in terms of hardware configurations and data distribution. Section \ref{sec:system_model} presents criteria for prioritizing FL clients and Section \ref{sec:implementation} provides implementation practices. The research challenges and opportunities are highlighted in Section \ref{sec:challenge} and Section \ref{sec:opportunity}, respectively.  
Last, this paper is concluded in Section \ref{sec:conclude}.

\section{Literature Review Process}
\label{sec:slr}

We adopt a systematic literature review process~\cite{slr1, slr2} in order to provide an unbiased and informative overview on FL client selection. 

\subsection{Research Questions}
\label{sec:research_question}
As an emerging topic in FL, client selection entails many unique design considerations. Thus, in this paper, we want to answer the following questions:
\begin{enumerate}
    \item How does FL clients behave that affects the client selection performance? (Section \ref{sec:heterogeneity})
    
    \item What is the principle behind existing FL client selection algorithms to prioritize FL clients? (Section \ref{sec:system_model})
    
    \item What is the current practice to implement an FL client selection algorithm? (Section \ref{sec:implementation})

    \item What are the challenges to realize an effective FL client selection algorithm? (Section \ref{sec:challenge})

    \item What are the research opportunities to boost performance of FL client selection? (Section \ref{sec:opportunity})
    
\end{enumerate}

\subsection{Literature Searching and Appraisal}

Measuring the utilities of FL clients and then scheduling clients based on their utility measures is the core idea behind FL client selection algorithms. Therefore, we search for works that have explicit utility measurements and scheduling decisions. \revise{We use the Scopus, Web of Science, and Google Scholar libraries and apply the following search syntax:}

\vspace{0.05in}
{
\centering
``Federated Learning'' AND \\
\centerline{
(``selection'' OR ``sampling'' OR ``scheduling'')
}
}
\vspace{0.05in}

\revise{
We sort the searching results of each library by relevance and use the first 20 papers of each library results as the starting point. We keep the papers that meet the following two criteria: (1) target the standard FL scenario (i.e., an FL server learns a model by collaborating with a bunch of FL clients) and (2) have explicit utility measurements and scheduling decisions. In other words, we do not consider papers that are designed for FL variants (e.g., hierarchical FL, serverless FL) or implicit machine learning scheduling (e.g., reinforcement learning-based client selection). Therefore, our identified representative works are general in principle and can be easily adapted to new designs.
Our initial literature searching results in 8 high-quality papers. Afterward, we read their citations and references, which includes 7 more high-quality papers. In the end, we identify \revise{15} papers as tabulated in Table \ref{tab:papers}.
}

\subsection{Literature Synthesis and Analysis}

For the selected papers, we carefully analyze the FL client behavior (Section~\ref{sec:heterogeneity}), categorize their methods to determine a client's priority (Section~\ref{sec:system_model}), summarize their implementation platforms (Section~\ref{sec:implementation}), identify the unsolved challenges (Section~\ref{sec:challenge}), and pinpoint the research opportunities (Section~\ref{sec:opportunity}). In the writing of this overview paper, we significantly integrate relevant background and works to provide a deep insight of FL client selection. 

\revise{

\subsection{Existing Surveys/Reviews}

\begin{table}[!t]
    \centering
    \centerline{
    \begin{tabular}{c| p{3in}}
        \hspace{0.0001in}\revise{Refs} & \hspace{0.0001in}\revise{Focus Point} \\
          \toprule
\revise{\cite{flsurvey20access}} & \hspace{0.0001in}\revise{Enabling software and hardware platforms, protocols, real-life applications and use-cases.} \\
\revise{\cite{flsurvey20communcationsurvey}} & \hspace{0.0001in}\revise{Communication costs, resource allocation, and privacy and security in the implementation of FL at scale.} \\
\revise{\cite{flsurvey21communcationsurvey2}} & \hspace{0.0001in}\revise{Sparsification, robustness, quantization, scalability, security, and privacy of FL-powered IoT applications.} \\
\revise{\cite{flsurvey21access}} & \hspace{0.0001in}\revise{Data partitioning, FL architectures, aggregation techniques, and personalization techniques.} \\
\revise{\cite{flsurvey21kbs}} & \hspace{0.0001in}\revise{Data partitioning, privacy mechanism, machine learning models, communication architecture and systems heterogeneity. }\\
\revise{\cite{flsurvey21iotj}} &  \hspace{0.0001in}\revise{Core system models and
designs, application areas, privacy and security, and resource
management.} \\ 
\revise{\cite{flsurvey21communcationsurvey}} & \hspace{0.0001in}\revise{IoT data sharing, data offloading and caching,
attack detection, localization, mobile crowdsensing, and IoT privacy and security.} \\
\revise{\cite{flsurvey22iotj}} & \hspace{0.0001in}\revise{Resource constrained IoT devices, distributed implementation, challenges and issues when applying FL to an IoT environment.} \\
\revise{\cite{flsurvey23knowledge}} & \hspace{0.0001in}\revise{Data distribution,  privacy mechanism, communication architecture, scale of federation and motivation of federation} \\
        \bottomrule
    \end{tabular}
    }
   \vspace{0.05in} \caption{\revise{Related surveys on FL, ordered by the publication year.}}
    \label{tab:surveys}
\end{table}

Due to the rapid development of federated learning, there are many published surveys on FL. Table~\ref{tab:surveys} tabulates several relevant surveys, which cover general aspects of FL from data distribution/partitioning to real-world implementations. However, these surveys only touch the topic of FL client selection without providing deep insights as this paper. Specifically, we adopt a literature review process to explain the most important aspects of FL client selection (see Section \ref{sec:research_question}) and thus readers can quickly obtain a comprehensive view of this research topic.

}

\section{FL Client Heterogeneity}
\label{sec:heterogeneity}

In a typical FL training scenario, FL clients exhibit system and statistical heterogeneity. 
Random client selection does not consider this heterogeneity and thus results in performance degradation. For example,
a large-scale study on real-world data from 136k smartphones shows that the heterogeneity reduces the FL model accuracy by up to 9.2\% and increases the convergence time by 2.64X~\cite{flash22tmc}.

\subsection{System Heterogeneity}

Mobile devices are usually equipped with different hardware that have diverse capabilities of computation, communication, energy, etc. 
\begin{itemize}
    \item 
    \emph{Computation Capability}.
    With the gaining popularity of gaming and AI applications,
    mobile devices often have AI accelerators such as GPU, NPU, or CUDA cores. However,  measurements of mainstream mobile devices show that they can spend more than tens of times difference in running AI models~\cite{aibenchmark}. The time difference is even large if AI models cannot fit into the memory of AI accelerators or AI model operators are not supported on mobile devices~\cite{mobile20icccn}. 
    
    \item \emph{Communication Capability}. Transmission speed is essential as FL training involves many rounds of model parameter transmissions between the server and the clients. 
    However, clients can have significantly different transmission speeds because of their transmission standards (e.g., LTE vs. WiFi), locations (indoor vs. outdoor), and wireless channel conditions (clean vs. congested). An analysis of hundreds of mobile phones in a real-world FL deployment~\cite{gboardquery} shows that the network bandwidth exhibits an order-of-magnitude difference~\cite{oort21osdi}. 
    
    \item \emph{Other Factors}. In addition to the computation and communication capabilities, many other factors also affect the availability and capability of clients. For example, a smartphone with a low battery level would automatically reduce the computation and communication capabilities to save energy; a mobile device with heavy applications running in the background greatly limits the available computing resources. 
    
\end{itemize}

\subsection{Statistical Heterogeneity}

Compared to other model training paradigms~(e.g., centralized machine learning~\cite{machine_learning}, conventional distributed machine learning~\cite{distributed_learning}), FL has unique data properties with regards to massively distributed data, unbalanced data, and non-Independent and Identically Distributed (IID) data~\cite{fedavg17aistats}.
\begin{itemize}
    \item \emph{Massively Distributed Data}: the number of FL clients is much larger than the clients' average number of data points. For example, a million of smartphones are involved in the Google keyboard query suggestion project~\cite{gboardquery}, but a user usually only makes up to dozens of queries a day. 
    
    \item \emph{Unbalanced Data}: clients have a different amount of data points. This is because the various use patterns result in a highly different local data size. For example, the Reddit comment dataset~\cite{reddit} reveals that 70\% users constitute the first quarter of the normalized number of comments while 10\% users make three-times more comments~\cite{oort21osdi}. 
    
    \item \emph{Non-IID Data}: each client's data does not represent the overall distribution as data are not IID. The non-IID data property has been widely observed in real-world applications~\cite{fedml20neurips}. Both the attribute skew and the label skew greatly affect the FL model training~\cite{non-iid21survey}.
\end{itemize}




\section{Prioritizing FL Clients}
\label{sec:system_model}

\begin{table}[!t]
    \centering
    \begin{tabular}{c l}
        Notation & Meaning  \\
 \toprule
        $N$ & Total number of FL clients \\
        $M$ & Number of clients selected in a training round \\ 
        $i$ & Index of FL clients \\
        $D_i$ & Data samples of client $i$ \\ 
        $B_i$ & A batch of data samples in client $i$ \\
        $f$ & Mapping from input to output \\
        $w$ & Model weights \\
        $j$ & Index of model weights \\ 
        $w_{ij}$, $\bar{{w}}_j$ & $j$-th weight in client $i$ and server, respectively \\
        $T$ & Deadline for an FL round \\
        $t_i$ & Round time of client $i$ \\
        $\alpha$ & Exponent controlling the punishment for stragglers \\
\bottomrule
    \end{tabular}
    \vspace{0.05in}
    \caption{Notations used in this paper.}
    \label{tab:notation}
\end{table}

In each training round, each client is measured for its utility/priority and the clients with the best utility measurements are selected for model training and aggregation. There are various methods for formulating clients' utility. Table \ref{tab:papers} summarizes the representative client selection algorithms that have explicit utility measurement and scheduling decisions. We tag the work designed specially for the system heterogeneity and the data heterogeneity. In this section, we elaborate on the utility functions. Table \ref{tab:notation} tabulates the notations that are used in this paper for quick reference.

An overall utility function for a client $i$ are often represented by its statistical utility and system utility as follows~\cite{oort21osdi, pyramidfl22mobicom}:

\begin{equation}
    Util(i) = Util_{stat}(i) \times Util_{sys}(i) ,
    \label{eq:system_model}
\end{equation}
where $Util_{stat}(i)$ and $Util_{sys}(i)$ represents the statistical and the system utility for client $i$, respectively.
Note that other factors, such as fairness, robustness, and privacy, can be formulated similarly. 

    \subsection{Statistical Utility}
\label{sec:statistical_utility}

Statistical utility represents the usefulness of a client's local update to the global model. We categorize statistical utilities into data sample-based and model-based. 


\subsubsection{Data Sample-Based Utility Measurement}

Data sample-based utility exploits a client's local data to quantify the statistical utility. 

\begin{itemize}

\item A simple way to represent the statistical utility of client $i$ is by the number of data samples on client $i$, i.e.,
\begin{equation}
    |D_i| ,
\end{equation}
where $D_i$ denotes the data samples of client $i$. 
This approach is valid when each data sample has the same quality, e.g., IID data. 

\item 
Importance sampling of data samples has been widely studied in the general ML literature~\cite{sample18icml,sample15icml} and has been recently applied in the federated setting~\cite{mercury2021sensys}. The idea is to assign a high importance score to a data sample that is divergent far from the model. Eq.~(\ref{eq:importance_sampling}) shows one optimal solution:
\begin{equation}
    |B_i| \sqrt{\frac{1}{|B_i|}\sum_{k\in B_i}|| \nabla f(k) ||_2^2} ,
    \label{eq:importance_sampling} 
\end{equation}
where $|| \nabla f(k) ||_2$ is the $L_2$-norm of the data sample $k$'s gradient in bin $B_i$ of FL client $i$. Albeit its optimality, the computation overhead is overwhelming as it needs to calculate the gradient of each data sample in all possible combinations. 
    
\item 
A data sample with a large loss generally has a large gradient norm~\cite{oort21osdi,nccb22aaai}. Therefore, an alternative to Eq.~(\ref{eq:importance_sampling}) is replacing the gradient norm with the loss, which results in 
\begin{equation}
 |B_i| \sqrt{\frac{1}{|B_i|}\sum_{k\in B_i}Loss(k)^2} .
\label{eq:oort_stat}
\end{equation}
Since each data sample's loss is available during the local training, the computation overhead is greatly reduced.    

\item 
The above expression can be further simplified by calculating the cumulative loss of client $i$'s data samples: 

\begin{equation}
    \sum_{k\in B_i}Loss(k) .
\end{equation}
It is adopted by \cite{powerofchoice22aistats} and also shows promising results.

\end{itemize}

\subsubsection{Model-Based Utility Measurement}

Another approach to determining the priority of a client is to compare its model weights/gradients. Different methods are developed to quantify the potential contribution of a local model based on the model. 

\begin{itemize}

\item 
The normalized model divergence is defined as the average difference between the model weights in client $i$ and the global model, i.e.,
\begin{equation}
    \frac{1}{|w|}\sum_{j=1}^{|w|}|\frac{w_{ij} - \bar{w_{j}}}{\bar{w_{j}}}|.
\end{equation}
$w$ represents the weights of a model and $\bar{w}$ represents the weights of the global model; $w_{ij}$ and 
$\bar{w_{j}}$ are the $j$-th weights of client $i$ and the global model, respectively. If the model divergence is small, then the local update from the client is insignificant and can be ignored~\cite{gaia17nsdi}.

\item 
Another work calculates the percentage of same-sign weights between a client model and the global model which can be regarded as a direction relative to the global model~\cite{cmfl19icdcs}:
\begin{equation}
    \frac{1}{|w|}\sum_{j=1}^{|w|}\mathbbm{1}(sign(w_{ij})=sign(\overset{-}{w_{j}})) ,
\end{equation}
where $\mathbbm{1}(sign(w_{ij})=sign(\overset{-}{w_{j}})) = 1$ if $w_{ij}$ and $\overset{-}{w_{j}}$ are of the same sign. 
Although most works believe that a more divergent local model is more important (e.g., \cite{oort21osdi,pyramidfl22mobicom,fcfl22imwut}),
\cite{cmfl19icdcs} shows that a lower percentage of same-sign weights results in better communication efficiency upon converges. 

\item The similarity to the convergence trend is also used to select clients with weights that are moving the most away from 0. For a L-layer model, the importance score of client $i$ can be expressed as~\cite{fcfl22imwut}: 
\begin{equation}
    \frac{1}{L} \sum_{l=1}^{L} \frac{mov(u_{il}) \cdot mov(\bar{u}_l)}{||mov(u_{il}) || \cdot || mov(\bar{u}_l) ||} ,
\end{equation}
where $u_{il}$ represents the gradient of the loss with respect to the weights of the $l$-th layer in client $i$. Correspondingly, $\bar{u}_l$ represents the counterpart of the global model. The movement function $mov(\cdot)$ represents the movement direction of the weights and is defined in~\cite{movepruning20neurips}.

\item 
Instead of comparing the local model with the global model, \cite{newt22transport} proposes to compare the change of local model before and after the local training. That is, 
\begin{equation}
     ||w_i^{after} - w_i^{before}||_2 ,
\end{equation}
A client $i$ is assumed to have a higher contribution if its local training results in a significant different local model \cite{newt22transport}.

\item 
Equivalently, the $L_2$-norm of the model's gradients can be used~\cite{wireless21twc, norm21icassp}, defined below. 

\begin{equation}
    || \nabla w_i ||_2 .
\end{equation}
A higher norm indicates a more valuable client. Variants of $L_2$-norm based client selection are also used, e.g., in \cite{norm20arxiv}.

\item 
The inner product between the gradients of the local model and the global represents its relative direction, which also indicates the divergence between a local model and the global model. 
\begin{equation}
     <\nabla w_{i}, \nabla \bar{w}> .
\end{equation}
FOLB~\cite{folb21jsac} and FedPNS~\cite{fedpns22tnse} removes clients that have negative inner products.

\end{itemize}

    \subsection{System Utility}
\label{sec:system_utility}

Due to the different hardware configurations, FL clients results in different system overheads (e.g., training and transmission time) in each training round. Slow devices (i.e., stragglers) can deteriorate the overall training performance by prolonging the training round if not carefully considered. There are a few mainstream system utilities to prioritize clients.  

\begin{itemize}

\item 
A deadline can be set to avoid excessively long server waiting time in each training round. The deadline-based selection has been widely used (e.g., \cite{gboard, fedcs19icc}), due to its easy implementation. Mathematically, clients with a deadline longer than $T$ are removed from the FL aggregation, i.e.,
\begin{equation}
    \mathbbm{1}(t_i < T) ,
\end{equation}
where $t_i$ is the total round time of client $i$ that includes the local training, transmission, compression, etc.  

\item 
A hard deadline, as above, might be too strict for some application scenarios. Thus, a soft deadline is imposed by some work (e.g., Oort~\cite{oort21osdi}) to
 penalize stragglers in the following manner:
    \begin{equation}
        (\frac{T}{t_i})^{\mathbbm{1}(T<t_i)\times \alpha} ,
    \label{eq:oort_sys}
    \end{equation}
where $\alpha$ is the exponent controlling the penalty for stragglers. 
Eq.~(\ref{eq:oort_sys}) equals 1 (i.e., no punishment) for non-stragglers and increases exponentially for stragglers.  
\end{itemize}

Note that $T$ can represent other 
metrics other than time. For example, if the system targets the computation speed, then $T$ is in FLOPs; if the system focuses the transmission bandwidth, then $T$ is in Mbps.
Many approaches set an empirical deadline, e.g., 2 minutes in the Gboard projects~\cite{gboard, gboardquery} to ignore straggler clients. Manually determining the deadline could result in inferior performance as $T$ significantly affects the clients for aggregation. FedBalancer designs an algorithm to automatically adjust $T$~\cite{fedbalancer22mobisys}, showing better performance than using the fixed deadline.

\subsection{Scheduling}
\label{sec:schedule}

Once the statistical utility function and the system utility function have been defined, the overall utility for each client can be calculated using Eq.~(\ref{eq:system_model}). Ideally, in each training round, each client is measured for its utility and the clients of the highest utility measures can be chosen to join the training. However, it is not practical to measure every client's utility in each training round, as 
a client’s utility often can only be determined after it has participated in a training round. Therefore, a mainstream approach is to forecast a client's utility along the training stage and update/rectify its utility measure once it is selected to join the training round~\cite{oort21osdi,pyramidfl22mobicom}. In addition, the scheduling is faced with the exploitation-exploration dilemma, which is explained in Section~\ref{sec:challenge}.

\subsection{Discussion}
\label{sec:remark}

In this paper, the overall utility of each client is a multiplication of the statistical utility and the system utility (Eq.~(\ref{eq:system_model})), the form of which has been adopted by recent works such as Oort~\cite{oort21osdi} and PyramidFL~\cite{pyramidfl22mobicom}.
The multiplication form can be extended to include other utility aspects such as fairness~\cite{qffl20iclr} and robustness~\cite{folb21jsac} by multiplying corresponding utility functions. It can also ignore unwanted aspects by simply assigning 1 to the corresponding utility functions. Therefore, the 
multiplication form is expressive.
Other forms are also applicable. For example, Eiffel~\cite{eiffel22tpds} expresses a client's overall utility by adding the loss value of its local model, the local data size, the computation power, the resource demand, and the age of update, with adjustable weights for different utility aspects. The utility functions covered in this section can be applied to other FL client selection designs. 

\section{Current Implementation}
\label{sec:implementation}

In this section, we briefly explain data simulation and frameworks for FL client selection research. 

\subsection{Data Simulation}

In research, FL clients' data are often assigned in one of the following manners. (1) Synthetic data partitions. Researchers can use  conventional ML datasets (e.g., MNIST~\cite{mnistdataset}, Shakespeare, CIFAR-10~\cite{cifar10}) and partition the dataset into different clients. This approach allows great flexibility as different degrees of data heterogeneity can be simulated~\cite{fedavg17aistats}. (2) Realistic data partitions. The other approach is to adopt datasets with the client ID and partition the dataset using the unique client ID. A variety of realistically partitioned dataset are available (refer to \cite{fedscale22icml, oort21osdi, fedml20neurips} for more information), such as OpenImage~\cite{openimage}, StackOverflow~\cite{stackoverflow}, and Reddit~\cite{reddit}. This data approach can more accurately capture the FL performance in real-world scenarios. 

\subsection{Federated Learning Frameworks}

In addition to implementing FL from scratch using e.g, PyTorch and TensorFlow, we can also use
frameworks that are designed specifically for FL, which could facilitate FL research and deployment. Below are some representative FL frameworks used for FL client selection research. 

\begin{itemize}

\item 
TensorFlow Federated (TFF)~\cite{tff} is developed by Google and is an open-source framework for experimenting with FL. It supports mobile devices and has been used for commercial projects such as mobile keyboard prediction~\cite{gboard}.

\item 
FedScale~\cite{fedscale22icml} is initialized by the University of Michigan. It provides many useful features, such as mobile device profiles of computation and communications speeds. FedScale enables on-device FL evaluation on smartphones and in-cluster simulations. Recent work such as  PyramidFL~\cite{pyramidfl22mobicom} and Oort~\cite{oort21osdi} use FedScale for their experiments.  
    
\item 
Leaf~\cite{leaf} is developed by a research group at Carnegie Mellon University. It includes a suite of federated datasets, an evaluation framework, and a set of reference implementations. Leaf has been used by FLASH~\cite{flash22tmc} to study the heterogeneity impacts on FL client selection algorithms. 
    
\item 
FedML~\cite{fedml20neurips} is an open-source platform for an end-to-end machine learning ecosystem. It supports a lightweight and cross-platform design for secure edge training. Therefore, FedML is useful for designing IoT-based FL systems~\cite{fediot21sensys}.
    
\end{itemize}

These representative FL frameworks have advantages and disadvantages. FedML focuses on FL ecosystems for diverse application domains, but it is still in its early stage with moderate community support. TFF is actively maintained by a large community. Building an industrial-level FL client selection solution with TFF is convenient as TFF can be easily integrated with other Google products/services. However, TFF does not provide dataset partitions designed specifically for FL research. To this end, Leaf is proposed to facilitate FL research by providing a unified API for several popular datasets but with no support for device profiles. As a newcomer to this field, FedScale is developed with both the statistical and system heterogeneity in mind from the beginning. However, its community support is still early and requires further testing.

\section{Challenges}
\label{sec:challenge}

In this section, we highlight several challenges that hinder the development of high-performance FL client selections. 

Existing work mostly assumes that devices are always available for FL training, which is not true in practice. Very often, devices are only available when they are idle, charged, and connected to WiFi, in order to protect the user experience~\cite{gboard}.
For example, Google observes lower training accuracy during the day, as few devices satisfy this requirement, which generally represents a skewed population~\cite{gboardquery}. 
Datasets for device availability are scarce. 
FLASH~\cite{flash22tmc} provides an input method App dataset containing smartphone status traces that can be used to emulate the device availability. Its data analysis reveals that some active devices dominate the global model, which leads to participant bias.

Always selecting the prioritized clients tends to result in sub-optimal performance as underrepresented clients may never have the chance to be selected~\cite{fcfl22imwut}. Therefore, there is a tradeoff between exploitation (selecting prioritized clients) and exploration (selecting more diverse clients). This exploitation-exploration dilemma is common in many research fields, such as data annotation in active learning~\cite{al20nc} and space search in reinforcement learning~\cite{rl21ijcai}. The exploitation-exploration dilemma is challenging, especially for FL, which needs more comprehensive studies of this problem. Current FL client selection work adopts simple methods to trade off the exploitation and exploration~\cite{oort21osdi, pyramidfl22mobicom}.

Designing an effective and general client selection algorithm remains challenging. Worse, heterogeneity is different across different regions and application scenarios. For example, mobile users in the US generally have stable network conditions, contrary to the widely assumed that mobile users always suffer from transmission interruptions~\cite{fcfl22imwut}. 

\section{Research Opportunities}
\label{sec:opportunity}

FL client selection is a new research topic with many unsolved problems and challenges. It also offers many research opportunities that are worth exploring. In addition to designing high-performance FL client selection algorithms for different application scenarios, the below aspects are also critical for FL client selection.

\subsection{Optimal Number of Selected Clients}

The current mainstream practice is to determine the number empirically. For example, the Google Gboard project uses 100 clients for training keyboard query suggestions. Many pieces of work have shown that the FL convergence rate can be improved by selecting more clients, with diminishing improvement gains as the number increases~\cite{fedavg17aistats}. Besides, more clients are preferred in each training round when data follow a more heterogeneous distribution~\cite{wireless21twc}. However, selecting more clients is not always possible or optimal when clients are subject to constraints such as energy or bandwidth~\cite{wireless21twc, ocean21twc}. 
Furthermore, different FL training stages may prefer different numbers of selected clients. A ``later-is-better'' phenomenon is observed in which an ascending number of client patterns is generally desired~\cite{ocean21twc}. However, all these observations are made in hindsight, and thus, research is needed to identify the optimal number of selected clients for diverse applications. Another promising line of research is automatically tuning the number during FL training. For example,  FedTune~\cite{fedtune22milcom} proposes a simple tuning algorithm that increases or decreases the number by one in each decision step. Overall, we need high-quality approaches designed specifically for tuning the number of selected clients in each training round. 

\subsection{Theoretical Performance Guarantee}

Existing work on FL client selection mostly rely on experiments to demonstrate their effectiveness; thus, the results given in these work may be susceptible to experiment bias.
For example, some work demonstrates that clients with more divergent models from the global are preferred (e.g., \cite{fcfl22imwut, oort21osdi}), while some work shows the opposite (e.g., \cite{cmfl19icdcs, fedpns22tnse}). The contradictory observations may stem from different application scenarios.
Due to the heuristic properties of most FL client selection algorithms, it is challenging, if possible, to provide theoretical guarantee for their algorithms with regard to model accuracy, convergence rate, robustness, fairness, etc. Without performance guarantee of client selection algorithms, FL practitioners tend to adopt the random client selection method in their projects, resulting in sub-optimal performance. Therefore, more research is needed to provide theoretical analysis frameworks for FL client selection algorithms.  

\subsection{Benchmark and Evaluation Metrics}

Existing work adopts various metrics to evaluate their performance. Final model accuracy~\cite{pyramidfl22mobicom}, time-to-accuracy~\cite{oort21osdi}, round-to-accuracy~\cite{fedNova20neurips}, transmission load~\cite{pa-cqls22wcnc}, etc, are well-known metrics. However, different metrics are not comparable. For example, selecting more clients in each training round results in a better round-to-accuracy performance. But it does not necessarily mean a better time-to-accuracy as the time length of each training round increases with the number of selected clients, nor a better transmission efficiency as more clients need to transmit model parameters in each training round. In addition, different experiment settings are adopted by existing work, whose results may not apply to other applications. \revise{Besides, the local data sampling of the selected clients can also affect the FL performance~\cite{fedss20icc}.} Therefore, the community needs well-established benchmarks and evaluation metrics to fairly and objectively compare different FL client selection algorithms. 

\revise{
\subsection{Extension to Other Federated Learning Scenarios}

In addition to the classical FL scenario, where a global model is trained using a bunch of clients (e.g., FedAvg settings), other variants of FL scenarios are gaining increasing attention from academia. For example, (1) Hierarchical FL~\cite{hierarchical22parallel}, which often includes cloud space, edge space, and client space compared to the standard server-client paradigm; (2) Cluster-based FL~\cite{haccs22ipdps}, which groups clients based on the data distribution or system capability into clusters and then schedules clusters for better performance; (3) Online FL~\cite{fleet22tist}, which requires lightweight FL updates during the user applications instead of the idle-charging-WiFi condition for client selection; (4) Serverless FL~\cite{cmfl22tpds}, where there is no fixed and permanent server for FL training coordination but in a peer-to-peer approach. Each variant poses unique challenges and entails specialized solutions. Therefore, more research efforts for client selection solutions are required to tackle different scenarios.
}

\subsection{Large-Scale Open FL Testbeds}

Although high-quality FL libraries are available for research and deployment, existing work mostly relies on either simulation or a small-scale implementation setting (e.g., a few devices). As a result, the observations made in existing work do not totally match the actual FL performance, especially for the FL client selection research, whose application scenarios often require a large number of clients. In addition, the current practice of building own experiment environments not only has difficulty in reproducing results but also hinders a fair comparison among different approaches. 
We envision large-scale, open testbeds for FL research, with a similar role to FlockLab testbed~\cite{flocklab2} for the wireless sensor network and IoT research.

\section{Conclusion}
\label{sec:conclude}

This paper is by no means an exhaustive survey, as FL client selection has many variants for different scenarios \revise{(e.g., trust-driven FL~\cite{trust22computing}, hybrid FL~\cite{hybrid22icc})}.
Also, we do not emphasize work that does not has explicit utility function and/or decision, e.g., fuzzy logic-based client selection~\cite{fuzzy22computer}, reinforcement learning-based client selection~\cite{multiarm20twc}. Instead, we cover the most general selection criteria that are widely applicable to new FL system designs. This paper provides insights into statistical and system heterogeneity, their utility functions, implementation suggestions, challenges, and research opportunities.
We hope this paper could inspire more research efforts in FL client selection.

\ifCLASSOPTIONcompsoc
  \section*{Acknowledgments}
\else
  \section*{Acknowledgment}
\fi

This work is partially supported by the Jiangsu Funding Program for Excellent Postdoctoral Talent (No. 2022ZB804).
This work is also supported by the National Science Foundation of the United States through grants USDA/NIFA 2020-67021-32855,  IIS-1838207, CNS 1901218, and OIA-2134901.  

\bibliographystyle{abbr_unsrt}
\bibliography{reference}

\begin{IEEEbiography}[\raisebox{0.15in}{\includegraphics[width=1.1in,clip,keepaspectratio]{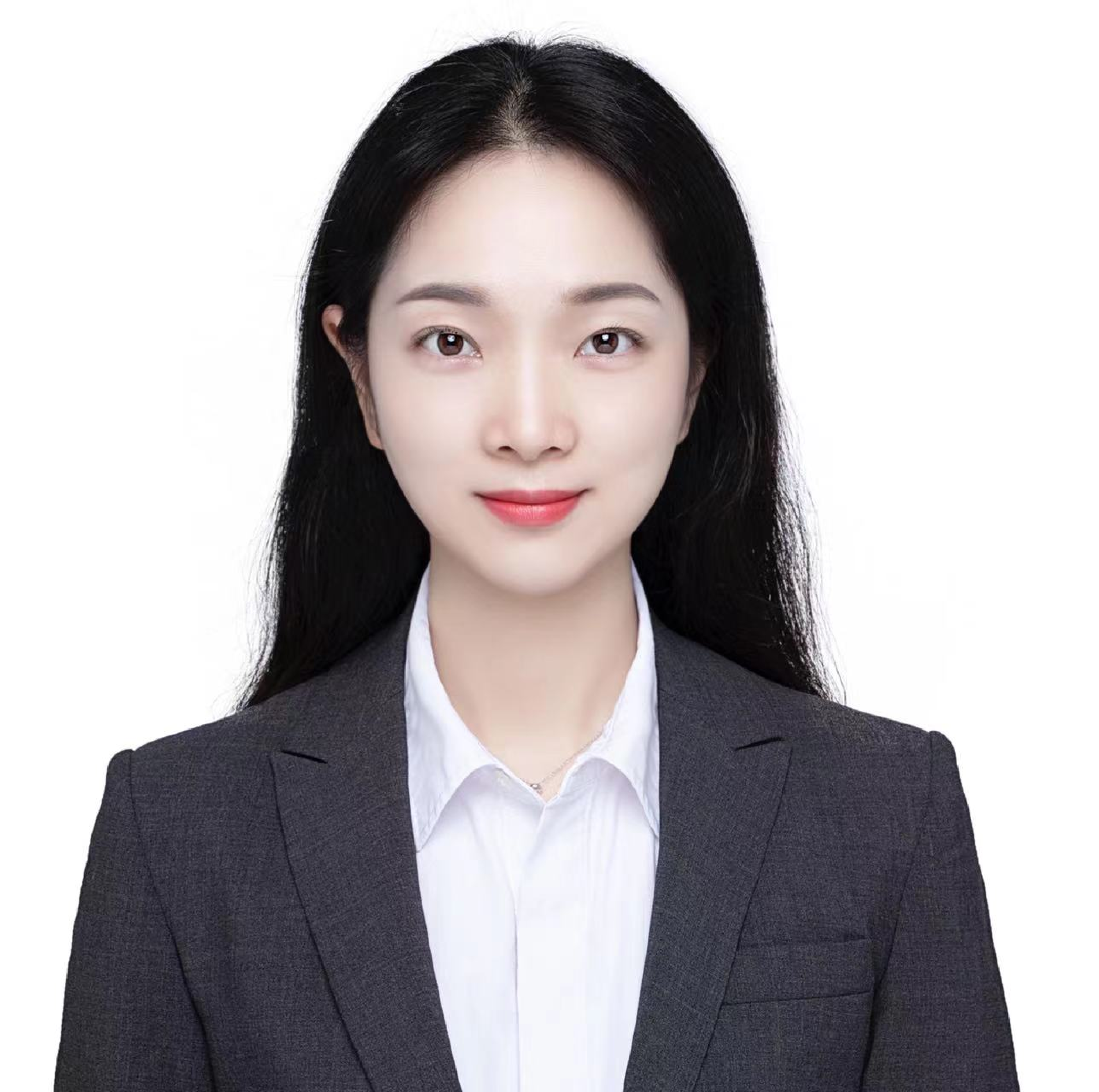}}]{Lei Fu} received her Ph.D. degree in Financial Information Engineering, Shanghai University of Finance and Economics in 2020. She is a postdoc in Fudan University and a quantitative investment analyst in Bank of Jiangsu, China. She received the Excellent Postdoc Award of Jiangsu Province in 2022. Her research interests include quantitative investment, 
macroeconomic computing, and financial technology. 
\end{IEEEbiography}

\begin{IEEEbiography}[{\includegraphics[width=1in,height=1.25in,clip,keepaspectratio]{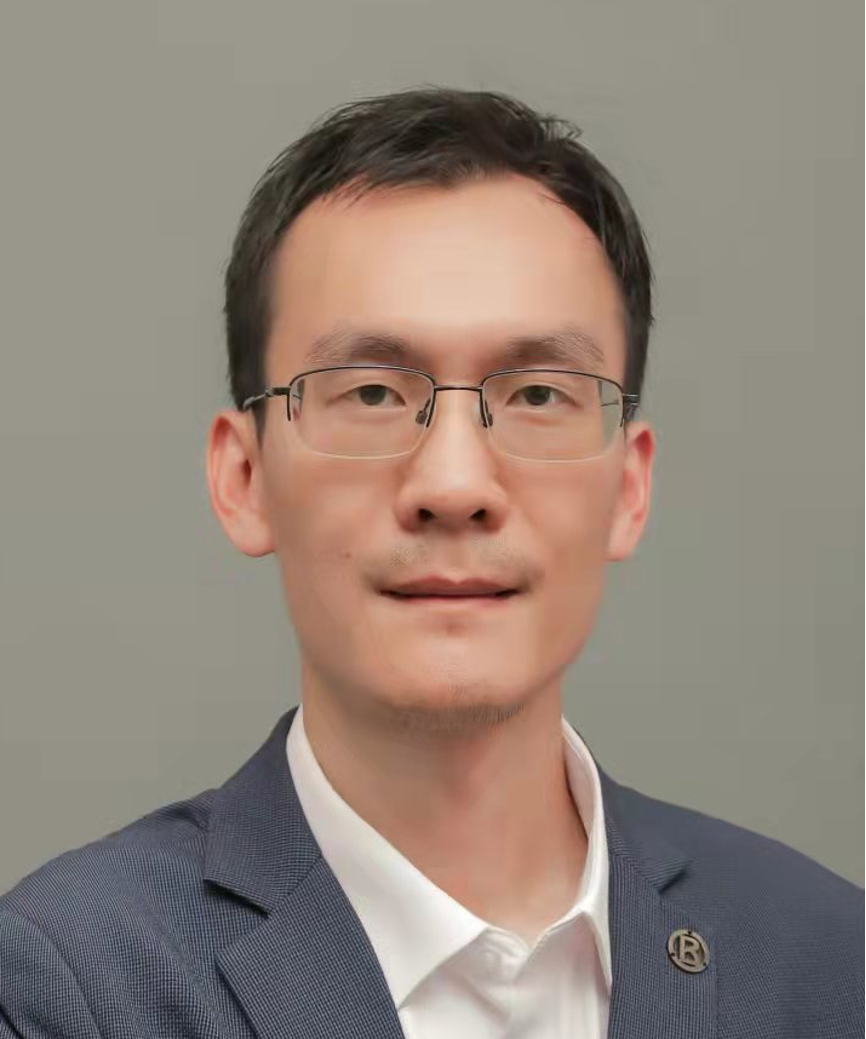}}]{Huanle Zhang}
is an associate professor in the School of Computer Science and Technology at Shandong University, China. He received his Ph.D. degree in Computer Science from the University of California, Davis (UC Davis), in 2020. He was employed as a postdoc at UC Davis 2020-2022 and a project officer at Nanyang Technological University 2014-2016. His research interests include data-centric AI, data privacy, IoT, and mobile systems. 
\end{IEEEbiography}

\begin{IEEEbiography}[\raisebox{0.4in}{\includegraphics[width=1in,height=1.25in,clip,keepaspectratio]{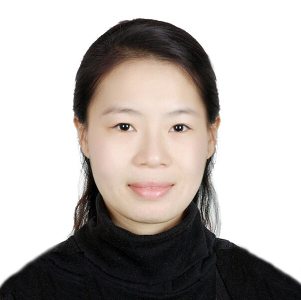}}]{Ge Gao} received her M.S. degree in Computer Science from Illinois Institute of Technology. Since 2020, she works at the School of Computer Science and Technology, Shandong University, China. Her current research interests include wearables activity data analysis, machine learning in healthcare, data privacy, and mobile computing.
\end{IEEEbiography}


\begin{IEEEbiography}[{\includegraphics[width=0.95in,clip,keepaspectratio]{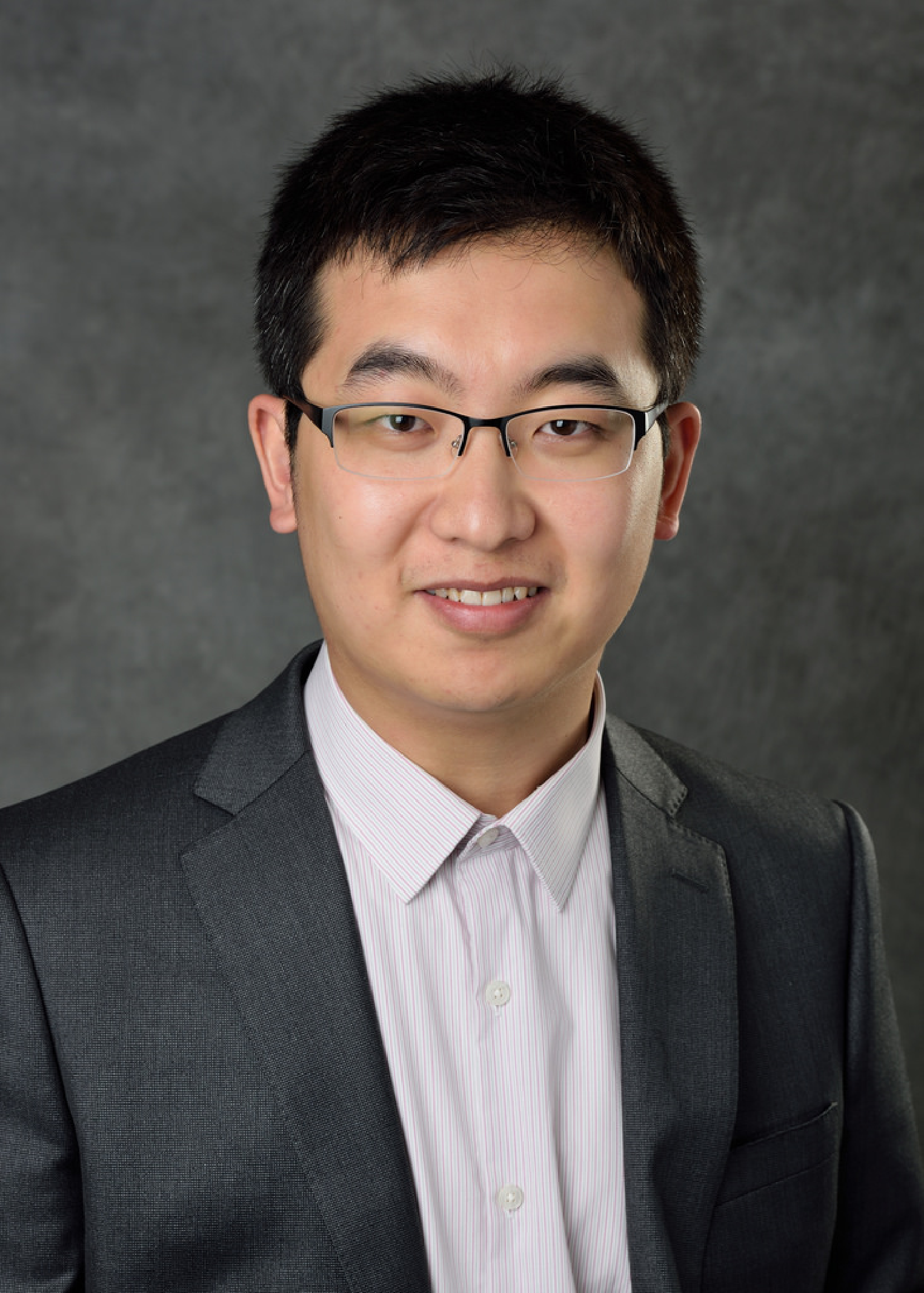}}]{Mi Zhang} is an associate professor in the Department of Computer Science and Engineering at The Ohio State University. He received his Ph.D. in Computer Engineering from the University of Southern California (USC) in 2013. His research lies at the intersection of mobile/edge/IoT systems and machine learning. He is the recipient of seven best paper awards and nominations. He is also the recipient of the National Science Foundation CRII Award, Facebook Faculty Research Award, Amazon Machine Learning Research Award, and MSU Innovation of the Year Award. He is a senior member of the IEEE.
\end{IEEEbiography}

\begin{IEEEbiography}[{\includegraphics[width=1.05in,clip,keepaspectratio]{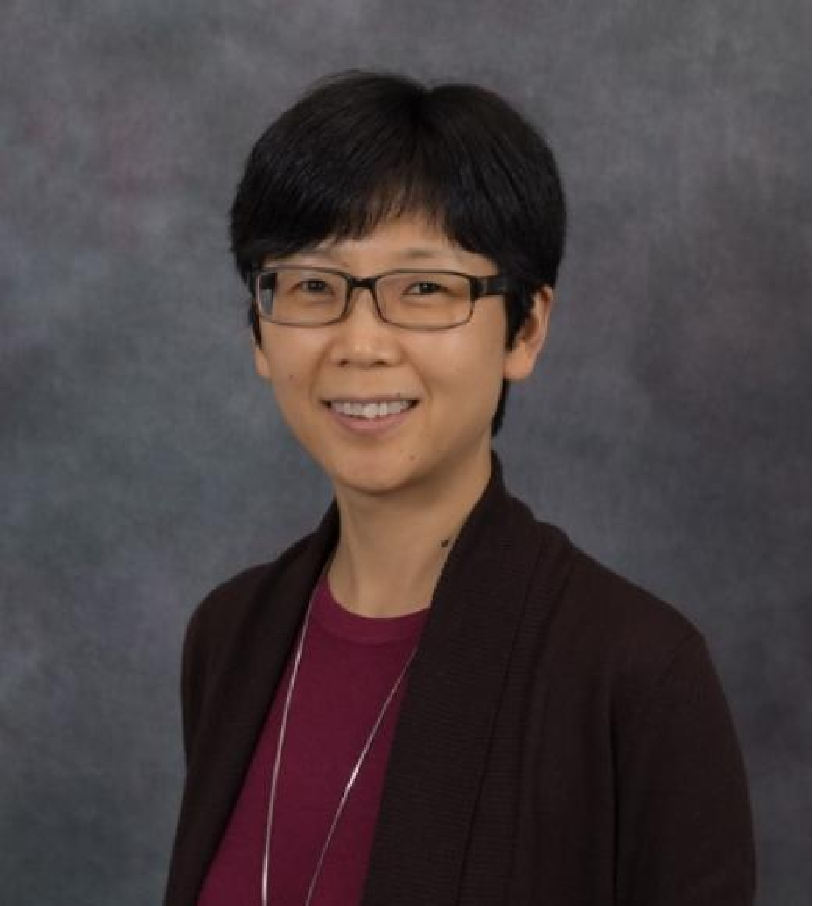}}]{Xin Liu} received her Ph.D. degree in electrical engineering from Purdue University in 2002. She is currently a Professor in Computer Science at the University of California, Davis. She received the Computer Networks Journal Best Paper of Year award in 2003 and NSF CAREER award in 2005. She became a Chancellor's Fellow in 2011. She is a co-PI and AI cluster co-lead for the USD 20M AI Institute for Next Generation Food Systems. She is a fellow of the IEEE.
\end{IEEEbiography}

 




\vfill

\end{document}